%% file: main.tex
\documentclass{article}

\usepackage{microtype}
\usepackage{graphicx}
\usepackage{subfigure}
\usepackage{booktabs} %

\usepackage{wrapfig}

\usepackage{hyperref}

\usepackage[accepted]{icml2023}

\usepackage{amsmath}
\usepackage{amssymb}
\usepackage{mathtools}
\usepackage{amsthm}

\usepackage[capitalize,noabbrev]{cleveref}

\theoremstyle{plain}

\theoremstyle{definition}

\theoremstyle{remark}

\usepackage{caption}

\usepackage[textsize=tiny]{todonotes}

\icmltitlerunning{BBF: Human-level Atari with human-level efficiency}

\begin{document}

\twocolumn[
\icmltitle{Bigger, Better, Faster: Human-level Atari with human-level efficiency}

\icmlsetsymbol{equal}{*}

\begin{icmlauthorlist}
\icmlauthor{Max Schwarzer}{equal,dm,mila,udem}
\icmlauthor{Johan Obando-Ceron}{equal,dm,mila,udem}
\icmlauthor{Aaron Courville}{mila,udem}
\icmlauthor{Marc G. Bellemare}{dm,mila,udem}
\icmlauthor{Rishabh Agarwal$^{\dagger}$}{dm,mila,udem}
\icmlauthor{Pablo Samuel Castro$^{\dagger}$}{dm}
\end{icmlauthorlist}

\icmlaffiliation{mila}{Mila}
\icmlaffiliation{udem}{Université de Montréal}
\icmlaffiliation{dm}{Google DeepMind}

\icmlcorrespondingauthor{Max Schwarzer}{MaxA.Schwarzer@gmail.com}
\icmlcorrespondingauthor{Johan Obando-Ceron}{jobando0730@gmail.com}

\icmlkeywords{Machine Learning, ICML}

\vskip 0.3in
]

\printAffiliationsAndNotice{\icmlEqualContribution} %

\def\alg{BBF}

\input{sections/abstract}

\input{sections/intro}

\input{sections/background}

\input{sections/relatedwork}

\input{sections/method}

\input{sections/analysis}

\input{sections/discussion}

\bibliography{references}
\bibliographystyle{icml2023}

\newpage
\appendix
\onecolumn

\paragraph{Acknowledgements.} Many thanks to Ross Goroshin, Georg Ostrovski, and Gopeshh Subbaraj for their feedback on an earlier draft of this paper.
The authors would like to thank the anonymous reviewers for useful discussions and feedback on this paper. We would also like to thank the Python community \cite{van1995python, 4160250}
for developing tools that enabled this work, including NumPy \cite{harris2020array}, Matplotlib \cite{hunter2007matplotlib} and JAX \cite{bradbury2018jax}.

\paragraph{Societal impact.} Although the work presented here is mostly academic, it aids in the development of more capable autonomous agents. While our contributions do not directly contribute to any negative societal impacts, we urge the community to consider these when building on our research.

\counterwithin{figure}{section}
\counterwithin{table}{section}
\counterwithin{equation}{section}
\input{sections/appendix}

\end{document}

%% file: sections/abstract.tex
\begin{abstract}
We introduce a value-based RL agent, which we call \alg, that achieves super-human performance in the Atari 100K benchmark. \alg\ relies on scaling the neural networks used for value estimation, as well as a number of other design choices that enable this scaling in a sample-efficient manner. We conduct extensive analyses of these design choices and provide insights for future work. We end with a discussion about updating the goalposts for sample-efficient RL research on the ALE.
\href{https://github.com/google-research/google-research/tree/master/bigger_better_faster}{\textbf{We make our code and data publicly available.}}
\end{abstract}

%% file: sections/intro.tex
\section{Introduction}
\label{sec:introduction}

Deep reinforcement learning~(RL) has been central to a number of successes including playing complex games at a human or super-human level, such as OpenAI Five~\citep{berner2019dota},  AlphaGo~\citep{silver16go}, and AlphaStar~\citep{vinyals2019grandmaster}, controlling nuclear fusion plasma in a tokomak \citep{degrave22magnetic}, and integrating human feedback for conversational agents \citep{ouyang2022training}. The success of these RL methods has relied on large neural networks and an enormous number of environment samples to learn from -- a human player would require tens of thousands of years of game play to gather the same amount of experience as OpenAI Five or AlphaGo. It is plausible that such large networks are necessary for the agent's value estimation and/or policy to be expressive enough for the environment's complexity, while large number of samples might be needed to gather enough experience so as to determine the long-term effect of different action choices as well as train such large networks effectively. As such,
obtaining human-level sample efficiency with deep RL remains an outstanding goal.

Although advances in modern hardware enable using large networks, in many environments it may be challenging to scale up the number of environment samples, especially for real-world domains such as healthcare or robotics. While approaches such as offline RL leverage existing datasets to reduce the need for environment samples \citep{agarwal2020optimistic}, the learned policies may be unable to handle distribution shifts when interacting with the real environment \citep{levine20offline} or may simply be limited in performance without online interactions \citep{ostrovski2021tandem}.

\begin{figure}[t!]
    \centering
    \includegraphics[width=1.05\linewidth]{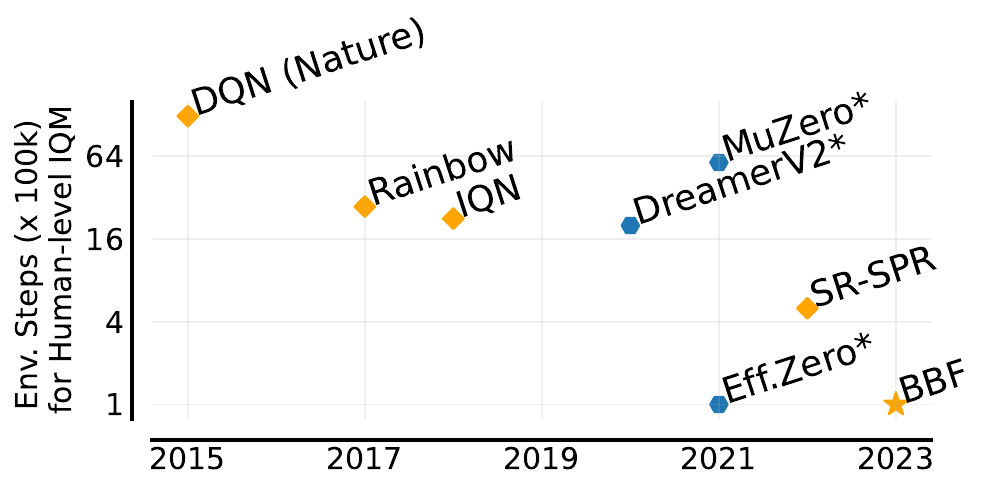}
    \vspace{-0.2cm}
    \caption{\textbf{Environment samples to reach human-level performance}, in terms of IQM~\citep{agarwal2021deep} over 26 games. Our proposed model-free agent, \alg, results in $5\times$ improvement over SR-SPR~\citep{doro2023sampleefficient} and at least $16\times$ improvement
    over representative model-free RL methods, including DQN~\citep{mnih2015human}, Rainbow~\citep{hessel2017rainbow} and IQN~\citep{dabney2018implicit}. To contrast with the sample-efficiency progress in model-based RL, we also include DreamerV2~\citep{hafner2020mastering}, MuZero Reanalyse~\citep{schrittwieser2021online} and EfficientZero~\citep{ye2021mastering}.} 
    \label{fig:sample_efficiency_progress}
    \vspace{-0.4cm}
\end{figure}

Thus, as RL continues to be used in increasingly challenging and sample-scarce scenarios, the need for scalable yet sample-efficient online RL methods becomes more pressing. Despite the variability in problem characteristics making a one-size-fits-all solution unrealistic, there are many insights that may transfer {\em across} problem domains. As such, methods that achieve ``state-of-the-art'' performance on established benchmarks can provide guidance and insights for others wishing to integrate their techniques.

\begin{figure*}[t]
    \vspace{0.1in}
    \centering
    \includegraphics[width=0.95\columnwidth]{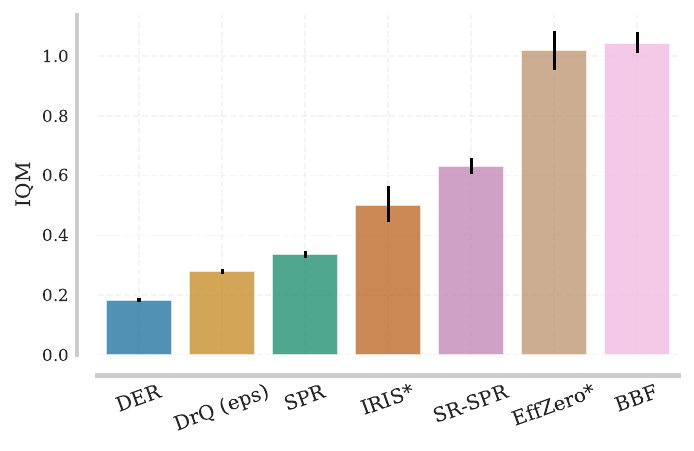}\hfill 
    \includegraphics[width=1.06\columnwidth]{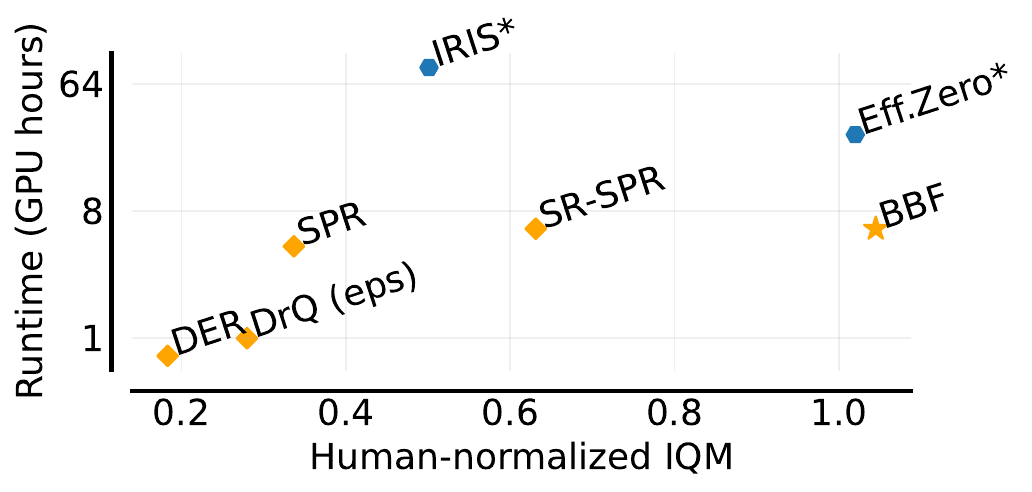}
    \caption{\textbf{Comparing Atari 100K performance and computational cost} of our model-free BBF agent to model-free SR-SPR~\citep{doro2023sampleefficient}, SPR~\citep{schwarzer2021dataefficient}, DrQ (eps)~\citep{kostrikov2020image} and DER~\citep{van2019use} as well as model-based$^*$ EfficientZero~\citep{ye2021mastering} and IRIS~\citep{micheli2023transformers}. \textbf{(Left)} BBF achieves higher performance than all competitors as measured by interquartile mean human-normalized over 26 games. Error bars show 95\% bootstrap CIs. \textbf{(Right)} Computational cost \emph{vs.} Performance, in terms of human-normalized IQM over 26 games. \alg\ results in $2\times$ improvement in performance over SR-SPR with nearly the same computational-cost, while results in similar performance to model-based EfficientZero with at least 4$\times$ reduction in runtime. For measuring runtime, we use the total number of A100 GPU hours spent per environment.
}
    \label{fig:toplinePerformance}
\end{figure*}

In this vein, we focus on the Atari 100K benchmark~\citep{Kaiser2020Model}, a well-known benchmark where agents are constrained to roughly 2 hours of game play, which is the amount of practice time the professional tester was given before human score evaluation. While human-level efficiency has been obtained by the model-based EfficientZero agent~\citep{ye2021mastering}, it has remained elusive for model-free RL agents. To this end, we introduce \alg, a model-free RL agent that achieves super-human performance -- interquartile mean~\citep{agarwal2021deep} human-normalized score above $1.0$ -- while being much more computationally efficient than EfficientZero~(Figure~\ref{fig:toplinePerformance}). Achieving this level of performance required a larger network than the decade-old 3-layer CNN architecture~\citep{mnih2013playing}, but as we will discuss below, scaling network size is not sufficient on its own. We discuss and analyze the various techniques and components that are necessary to train \alg\ successfully and provide guidance for future work to build on our findings. Finally, we propose moving the goalpost for sample-efficient RL research on the ALE.

%% file: sections/background.tex
\section{Background}
\label{sec:background}

\begin{figure*}[t]
    \centering
    \includegraphics[width=0.98\textwidth]{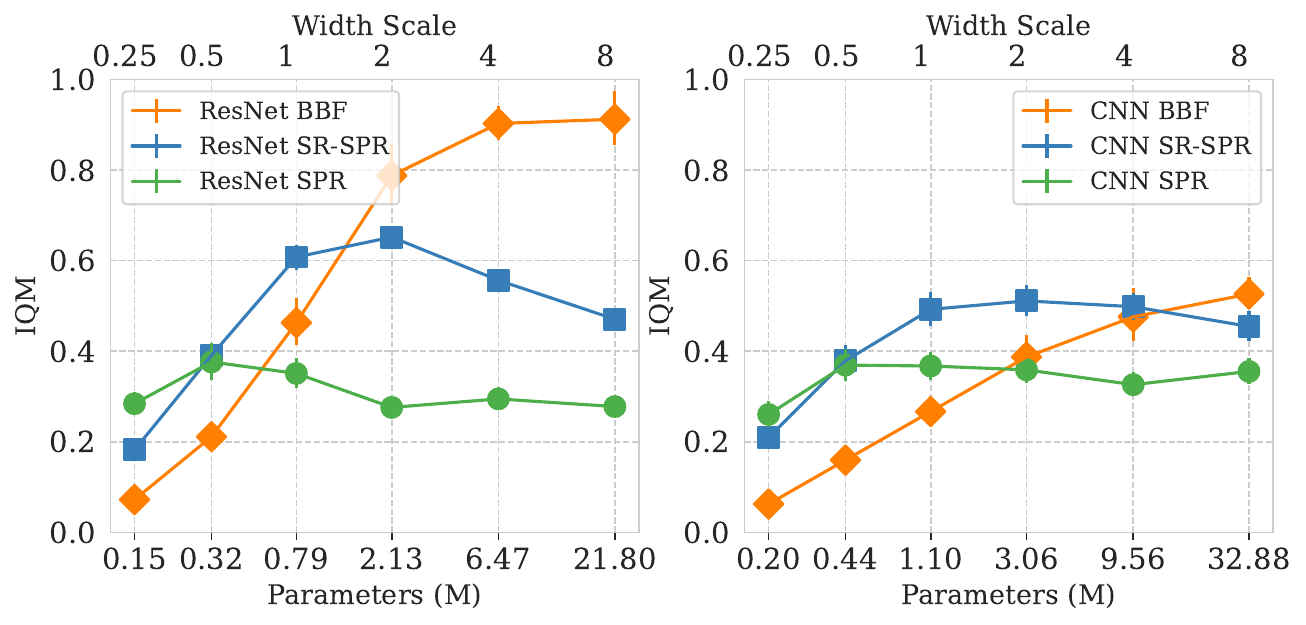}
    \vspace{-0.1cm}
    \caption{\textbf{Scaling network widths for both ResNet and CNN architectures}, for BBF, SR-SPR and SPR at replay ratio 2, with an Impala-based ResNet \textbf{(left)} and the standard 3-layer CNN~\citep{mnih2015human} \textbf{(right)}. We report interquantile mean performance with error bars indicating 95\% confidence intervals. On the x-axis we report the approximate parameter count of each configuration as well as its width relative to the default (width scale = 1).}
    \label{fig:scalingWidths}
    \vspace{-0.2cm}
\end{figure*}

The RL problem is generally described as a Markov Decision Proces (MDP) \citep{puterman2014markov}, defined by the tuple $\langle \mathcal{S}, \mathcal{A}, \mathcal{P}, \mathcal{R} \rangle$, where $\mathcal{S}$ is the set of states, $\mathcal{A}$ is the set of available actions, $\mathcal{P}:\mathcal{S}\times\mathcal{A}\rightarrow \Delta(\mathcal{S})$\footnote{$\Delta(S)$ denotes a distribution over the set $S$.} is the transition function, and $\mathcal{R}:\mathcal{S}\times\mathcal{A}\rightarrow\mathbb{R}$ is the reward function.
Agent behavior in RL can be formalized by a policy $\pi:\mathcal{S}\rightarrow\Delta(\mathcal{A})$, which maps states to a distribution of actions. The {\em value} of $\pi$ when starting from $s\in\mathcal{S}$ is defined as the discounted sum of expected rewards: $V^{\pi}(s) := \mathbb{E}_{\pi, \mathcal{P}}\left[\sum_{t=0}^{\infty} \gamma^t r\left(s_t, a_t\right)\right]$, where $\gamma\in [0, 1)$ is a discount factor that encourages the agent to accumulate rewards sooner rather than later. The goal of an RL agent is to find a policy $\pi^*$ that maximizes this sum: $V^{\pi^*} \geq V^{\pi}$ for all $\pi$.

\sloppy While there are a number of valid approaches \citep{sutton98rl}, in this paper we focus on model-free {\em value-based} methods. Common value-based algorithms approximate the $Q^*$-values, defined via the Bellman recurrence:\\ $Q^*(s, a) :=  R(s, a) + \gamma \mathbb{E}_{s'\sim\mathcal{P}(s, a)}[\max_{a'\in\mathcal{A}}Q^*(s', a')]$. The optimal policy $\pi^*$ can then be obtained from the optimal state-action value function $Q^*$ as $\pi^*(x) := \max_{a\in \mathcal{A}}Q^*(s, a)$.
A common approach for learning $Q^*$ is the method of temporal differences, optimizing the Bellman temporal difference: 
$$\left(r\left( {s}_{t}, {a}_t\right)+\gamma \max _{{a}_{t+1}} Q\left({s}_{t+1}, {a}_{t+1}\right)\right)-Q\left({s}_t, {a}_t\right).$$ We often refer to $\left(r\left({s}_t, {a}_t\right)+\gamma \max _{{a}_{t+1}} Q\left({s}_{t+1}, {a}_{t+1}\right)\right)$ as the {\em Bellman target}.

\citet{mnih2015humanlevel} introduced the agent DQN by combining temporal-difference learning with deep networks, and demonstrated its capabilities in achieving human-level performance on the Arcade Learning Environment (ALE) \citep{bellemare2012ale}. They used a network consisting of 3 convolutional layers and 2 fully connected layers, parameterized by $\theta$, to approximate $Q$ (denoted as $Q_{\theta}$). We will refer to this architecture as the CNN architecture.
Most of the work in value-based agents is built on the original DQN agent, and we discuss a few of these advances below which are relevant to our work.

\citet{Hessel2018RainbowCI} combined six components into a single agent they called {\em Rainbow}: prioritized experience \citep{Schaul2016PrioritizedER}, $n$-step learning \citep{sutton88learning}, distributional RL \citep{Bellemare2017ADP}, 
double Q-learning~\citep{hasselt2015doubledqn}, dueling architecture~\citep{wang16dueling} and NoisyNets~\citep{fortunato18noisy}. \citet{Hessel2018RainbowCI} and \citet{ceron2021revisiting} both showed that Multi-step learning is one of the most crucial components of Rainbow, in that removing it caused a large drop in performance.%

In $n$-step learning, instead of computing the temporal difference error using a single-step transition, one can use $n$-step targets instead \cite{sutton88learning}, where for a trajectory $(s_0, a_0, r_0, s_1, a_1, \cdots)$ and update horizon $n$: $R_t^{(n)} := \sum_{k=0}^{n-1}\gamma^k r_{t+k+1}$, yielding the multi-step temporal difference: $R_t^{(n)} + \gamma^n\max_{a'}Q_{{\theta}}(s_{t+n}, a') - Q_{\theta}(s_t, a_t)$.

Most modern RL algorithms store past experiences in a {\em replay buffer} that increases sample efficiency by allowing the agent to use samples multiple times during learning, and to leverage modern hardware such as GPUs and TPUs by training on sampled mini-batches. An important design parameter is the \textbf{replay ratio}, the ratio of learning updates to online experience collected \citep{fedus2020revisiting}. It is worth noting that DQN uses a replay ratio of $0.25$ (4 environment interactions for every learning update), while some sample-efficient agents based on Rainbow use a value of $1$. %

\citet{nikishin22primacy} showed that the networks used by deep RL agents have a tendency to overfit to early experience, which can result in sub-optimal performance. They proposed a simple strategy consisting of periodically resetting the parameters of the final layers of DQN-based agents to counteract this. Building on this promising work, \citet{doro2023sampleefficient} added a shrink-and-perturb technique for the parameters of the convolutional layers, and showed that this allowed them to scale the replay ratio to values as high as 16, with no performance degradation.

%% file: sections/relatedwork.tex
\begin{figure*}[ht!]
    \centering
    \includegraphics[width=0.98\columnwidth]{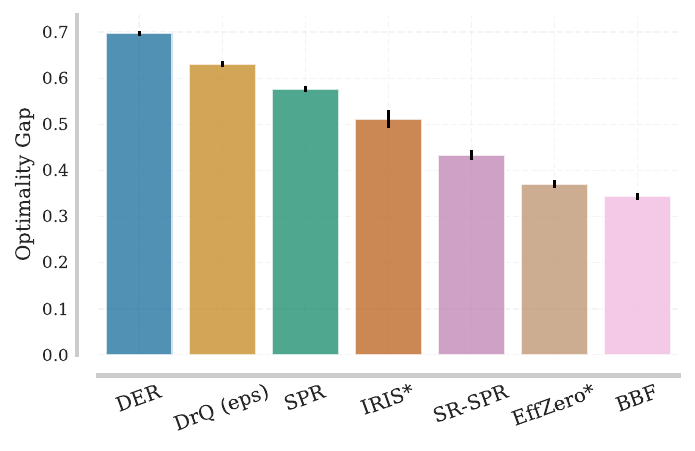}\hfill
    \includegraphics[width=\columnwidth]{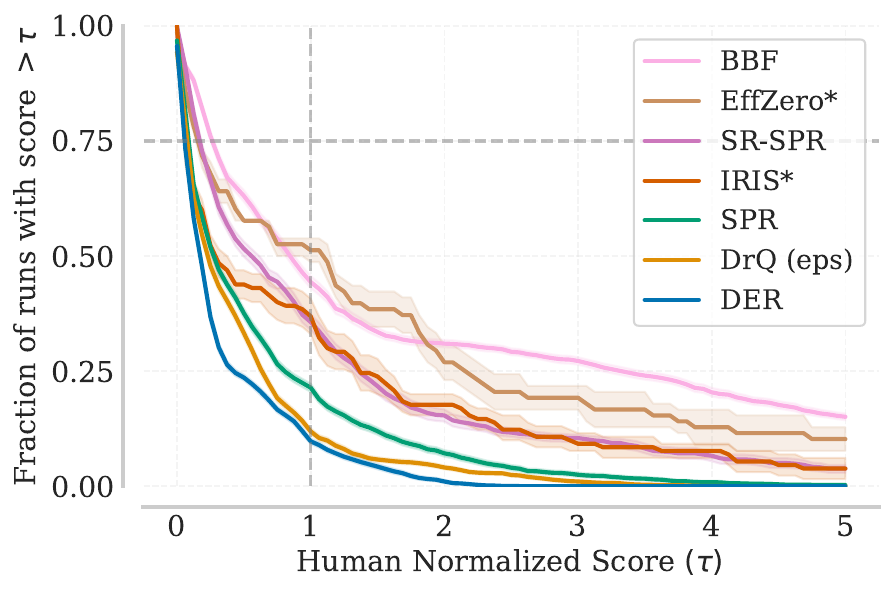}
    \caption{(\textbf{Left}). Optimality Gap (lower is better) for BBF at replay ratio 8 and competing methods on Atari 100K. Error bars show 95\% CIs. BBF, has a lower optimality gap than any competing algorithm, indicating that it comes closer on average to achieving human-level performance across all tasks. \textbf{(Right)}
     Performance profiles showing the distribution of scores across all runs and 26 games at the end
of training (higher is better). Area under an algorithm’s profile is its mean performance while $\tau$ value where it
intersects y = 0.75 shows its 25th percentile performance. BBF has better performance on challenging tasks that may not otherwise contribute to IQM or median performance.}
    \label{fig:optimalitygap}
\end{figure*}

\section{Related Work}
\label{sec:related_work}

\noindent {\bf Sample-Efficient RL on ALE:} Sample efficiency has always been an import aspect of evaluation in RL, as it can often be expensive to interact with an environment. \citet{Kaiser2020Model} introduced the Atari 100K benchmark, which has proven to be useful for evaluating sample-efficiency, and has led to a number of recent advances. %

\citet{kostrikov2020image} use data augmentation to design a sample-efficient RL method, DrQ, which outperformed prior methods on Atari 100K. %
Data-Efficient Rainbow (DER)~\citep{van2019use} and DrQ($\epsilon$) \citep{agarwal2021deep} simply modified the hyperparameters of existing model-free algorithms to exceed the performance of existing methods %
without any algorithmic innovation.

\citet{schwarzer2021dataefficient} introduced SPR, which builds on Rainbow~\citep{hessel2017rainbow} and uses a self-supervised temporal consistency loss based on BYOL~\citep{grill2020bootstrap} combined with data augmentation. SR-SPR~\citep{schwarzer2021dataefficient} combines SPR with periodic network resets to achieve state-of-the-art performance on the 100K benchmark. %
\citet{ye2021mastering} used a self-supervised consistency loss similar to SPR~\citep{chen2021exploring}. 

EfficientZero \citep{ye2021mastering},  an efficient variant of MuZero \citep{schrittwieser2020mastering}, learns a discrete-action latent dynamics model from environment interactions, and selects actions via lookahead MCTS %
in the latent space of the model. \citet{micheli2023transformers} introduce IRIS, a data-efficient agent that learns in a world model composed of an autoencoder and an auto-regressive Transformer.

\noindent {\bf Scaling in Deep RL:} Deep neural networks are useful for extracting features from data
relevant for various downstream tasks. Recently, there has been interest in the scaling properties of neural network architectures, as scaling model size has led to commensurate performance gains in applications ranging from language modelling to computer vision. 

Based on those promising gains, the deep RL community has begun to investigate the effect of increasing the model size of the function approximator. \citet{sinha2020d2rl} and \citet{ota2021training} explore the interplay between the size, structure, and performance of deep RL agents to provide intuition and guidelines for using larger networks.
\citet{kumar_scales2023} find that with ResNets~(up to 80 million parameter networks) combined with distributional RL and feature normalization, offline RL can exhibit strong performance that scales with model capacity. %
\citet{taiga2023investigating} show that generalization capabilities on the ALE benefit from higher capacity networks, such as ResNets. %
\citet{cobbe2020leveraging} and  \citet{farebrother2023protovalue} demonstrate benefits when scaling the number of features in each layer of the ResNet architecture used by Impala~\citep{espeholt2018impala}, which motivated the choice of feature width scaling in this work. Different from these works,  our work focus on improving sample-efficiency in RL as opposed to offline RL or improving generalization in RL. 

In the context of online RL, \citet{danijar_dreamerv3} demonstrate that increased dynamics model size, trained via supervised learning objectives, leads to monotonic improvements in the agent's final performance. Recently, AdA~\citep{team2023human} scales transformer encoder for a Muesli agent up to 265M parameters. Interestingly, AdA required distillation from smaller models to bigger models to achieve this scaling, in the spirit of reincarnating RL~\citep{agarwalreincarnating}. However, it is unclear whether findings from above papers generalize to scaling typical value-based deep RL methods in sample-constraint settings, which we study in this work.

%% file: sections/method.tex
\section{Method}
\label{sec:method}

\begin{figure*}[!t]
    \centering
    \includegraphics[width=0.9\textwidth]{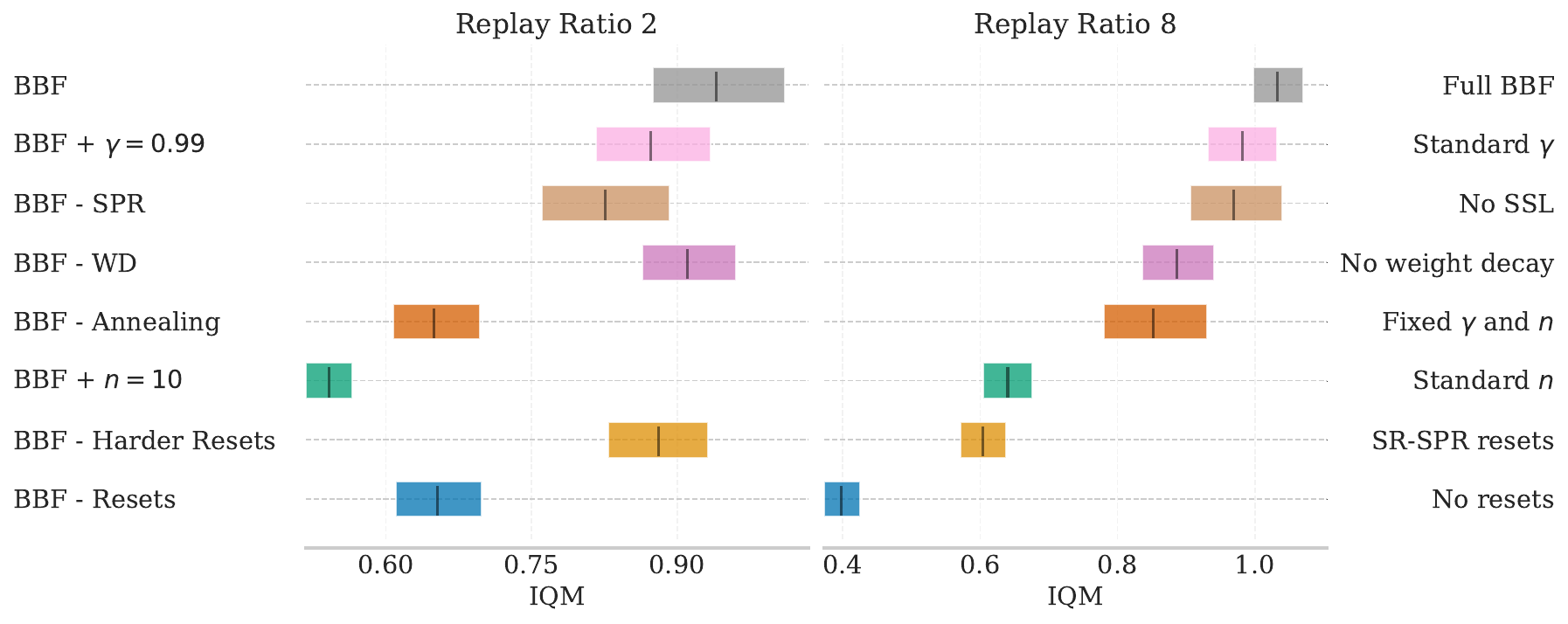}
    \caption{\textbf{Evaluating the impact of removing the various components that make up BBF with RR=2 and RR=8}. Reporting interquantile mean averaged over the 26 Atari 100k games, with 95\% CIs over 15 independent runs.}
    \label{fig:bbfAblations}
    \vspace{-0.15cm}
\end{figure*}

The question driving this work is: {\em How does one scale networks for deep RL when samples are scarce?} To investigate this, we focus on the well-known Atari 100K benchmark \citep{Kaiser2020Model}, which includes 26 Atari 2600 games of diverse characteristics, where the agent may perform only 100K environment steps, roughly equivalent to two hours of human gameplay\footnote{100k steps (400k frames) at 60 FPS is 111 minutes.}. As we will see, n{\"a}ively scaling networks can rarely maintain performance, let alone improve it.

The culmination of our investigation is the Bigger, Better, Faster agent, or BBF in short, which achieves super-human performance on Atari 100K with about 6 hours on single GPU. \autoref{fig:toplinePerformance} demonstrates the strong performance of BBF relative to some of the best-performing Atari 100K agents: EfficientZero \citep{ye2021mastering}, SR-SPR \citep{doro2023sampleefficient}, and IRIS \citep{micheli2023transformers}. BBF consists of a number of components, which we discuss in detail below.

Our implementation is based on the Dopamine framework \citep{castro18dopamine} and uses mostly already previously-released components. For evaluation, we use rliable~\citep{agarwal2021deep} and in particular, the interquartile mean~(IQM) metric, which is the average score of the middle 50\% runs combined across all games and seeds.

\paragraph{Base agent.} BBF uses a modified version of the recently introduced SR-SPR agent \citep{doro2023sampleefficient}. Through the use of periodic network resets, SR-SPR is able to scale up its replay ratio (RR) to values as high as 16, yielding better sample efficiency. For BBF, we use RR=8 in order to balance the increased computation arising from our large network. Note that this is still very high relative to existing Atari agents -- Rainbow and its data-efficient variant DER~\citep{van2019use} use RR=0.25 and 1, respectively.

As we expect that many users will not wish to pay the computational costs of running at replay ratio 8, we also present results for BBF and ablations at replay ratio 2 (matching SPR). For all experiments we state which replay ratio is being used in the captions.

\paragraph{Harder resets.} The original SR-SPR agent~\citep{doro2023sampleefficient} used a shrink-and-perturb method for the convolutional layers where parameters were only perturbed 20\% of the way towards a random target, while later layers were fully reset to a random initialization. An interesting result of our investigation is that using harder resets of the convolutional layers yields better performance. In our work, we move them 50\% towards the random target, resulting in a stronger perturbation and improving results (see \autoref{fig:bbfAblations}). This may be because larger networks need more regularization, as we find that they reduce loss faster (\autoref{fig:learning_curves}).

\paragraph{Larger network.} Scaling network capacity is one of the motivating factors for our work. As such, we adopt the Impala-CNN~\citep{espeholt2018impala} network, a 15-layer ResNet, which has previously led to substantial performance gains over the standard 3-layer CNN architecture in Atari tasks where large amounts of data are available~\citep{agarwalreincarnating,schmidt2021fast}. Additionally, BBF scales the width of each layer in Impala-CNN by 4$\times$. In \autoref{fig:scalingWidths}, we examine how the performance of SPR, SR-SPR and BBF varies with different choices of scaling width, for both the ResNet and original CNN architectures. Interestingly, although the CNN has roughly 50\% more parameters than the ResNet at each scale level, the ResNet yields better performance at all scaling levels for both SR-SPR and BBF.

What stands out from \autoref{fig:scalingWidths} is that BBF's performance continues to grow as width is increased, whereas SR-SPR seems to peak at 1-2$\times$ (for both architectures). Given that ResNet BBF performs comparably at 4$\times$ and 8$\times$, we chose 4$\times$ to reduce the computational burden. While reducing widths beyond this could further reduce computational costs, this comes at the cost of increasingly sharp reductions in performance for all methods tested.

\begin{figure}[!t]
    \centering
    \includegraphics[width=\columnwidth]{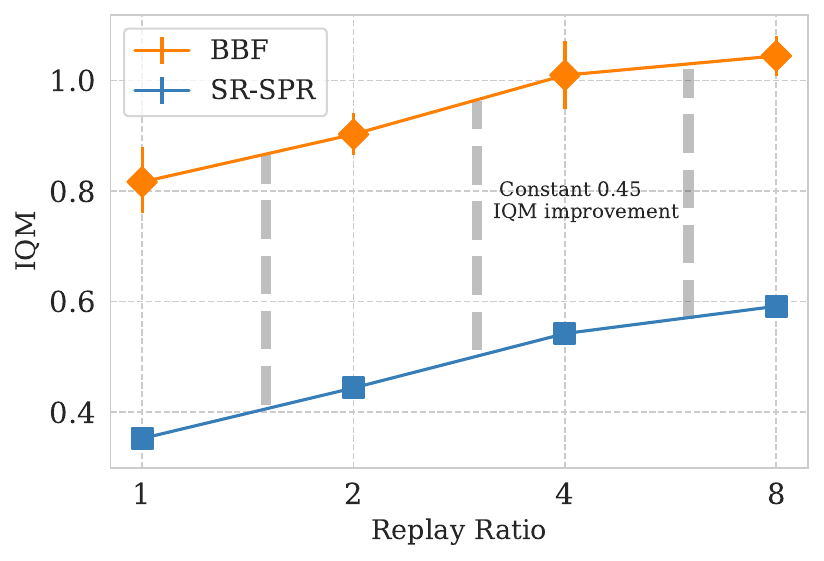}
    \vspace{-0.5cm}
    \caption{\textbf{Comparison of BBF and SR-SPR across different replay ratios}. We report IQM with 95\% CIs for each point. BBF achieves an almost-constant 0.45 IQM improvement over SR-SPR at each replay ratio.}
    \label{fig:replayRatioComparison}
    \vspace{-0.5cm}
\end{figure}

\paragraph{Receding update horizon.} One of the surprising components of BBF is the use of an update horizon~($n$-step) that decreases exponentially from 10 to 3 over the first 10K gradient steps following each network reset. Given that we follow the schedule of \citet{doro2023sampleefficient} and reset every 40k gradient steps, the annealing phase is always 25\% of training, regardless of the replay ratio. As can be seen in \autoref{fig:bbfAblations}, this yields a much stronger agent than using a fixed value of $n=3$, which is default for Rainbow, or $n=10$, which is typically used by Atari 100K agents like SR-SPR.

Our $n$-step schedule is motivated by the theoretical results of \citet{kearns2000bias} -- larger values of $n$-step leads to faster convergence but to higher asymptotic errors with respect to the optimal value function.  Thus, selecting a fixed value of $n$ corresponds to a choice between having either rapid convergence to a worse asymptote, or slower convergence to a better asymptote. As such, our exponential annealing schedule closely resembles the optimal decreasing schedule for $n$-step derived by \citet{kearns2000bias}.

\paragraph{Increasing discount factor.} Motivated by findings that increasing the discount factor $\gamma$ during learning improves performance~\citep{franccois2015discount}, we increase $\gamma$ from $\gamma_1$ to $\gamma_2$, following the same exponential schedule as for the update horizon. Note that increasing $\gamma$ has the effect of progressively giving more weights to delayed rewards. We choose $\gamma_1=0.97$, slightly lower than the typical discount used for Atari, and $\gamma_2=0.997$ as it is used by MuZero~\citep{schrittwieser2021online} and EfficientZero~\citep{ye2021mastering}. As with the update horizon, \autoref{fig:bbfAblations} demonstrates that this strategy outperforms using a fixed value. 

\begin{figure}
    \centering
    \includegraphics[width=\columnwidth]{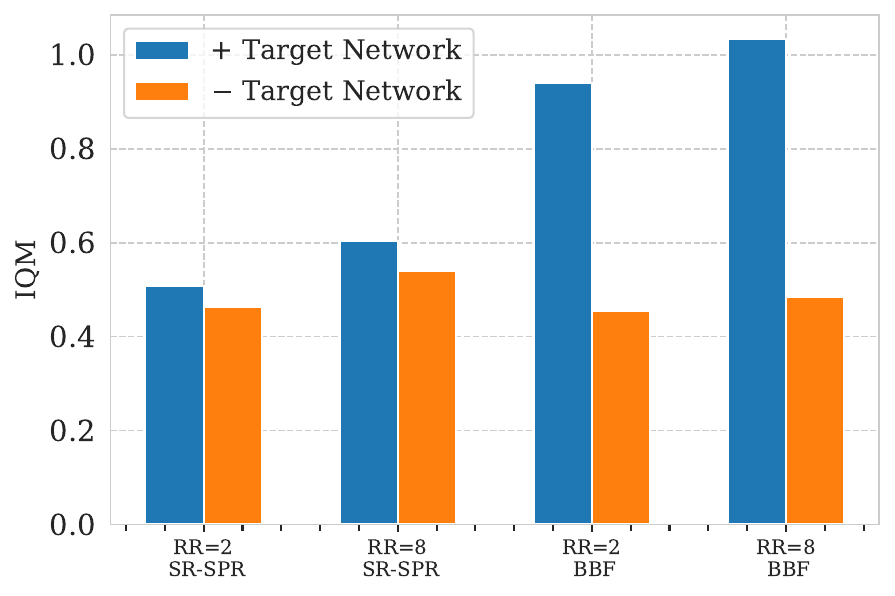}
    \vspace{-0.5cm}
    \caption{\textbf{Comparison of BBF and SR-SPR at replay ratios 2 and 8 with and without EMA target networks}. Human-normalized IQM on the 26 Atari 100k games.}
    \label{fig:target_figure}
    \vspace{-0.5cm}
\end{figure}

\paragraph{Weight decay.} We incorporate weight decay in our agent to curb statistical overfitting, as BBF is likely to overfit with its high replay ratio. To do so, we use the AdamW optimizer~\citep{loshchilov2018decoupled} with a weight decay value of $0.1$. \autoref{fig:bbfAblations} suggests the gains from adding weight decay are significant and increase with replay ratio, indicating that the regularizing effects of weight decay enhance replay ratio scaling with large networks.

\paragraph{Removing noisy nets.}  Finally, we found that NoisyNets \citep{fortunato2018noisy}, used in the original SPR  \citep{schwarzer2021dataefficient} and SR-SPR, did not improve performance. This could be due to NoisyNets causing over-exploration due to increased policy churn~\citep{schaul2022phenomenon} from added noise during training, or due to added variance in optimization, and we leave investigation to future work. Removing NoisyNets results in large computational and memory savings, as NoisyNets creates duplicate copies of the weight matrices for the final two linear layers in the network, which contain the vast majority of all parameters: turning on NoisyNets increases the FLOPs per forward pass and the memory footprint by a factor of 2.5$\times$ and 1.6$\times$, respectively, which both increases runtime and reduces the number of training runs that can be run in parallel on a single GPU. Removing NoisyNets is thus critical to allowing BBF to achieve reasonable compute efficiency despite its larger networks. We found that this decision had no significant impact on task performance (see \autoref{fig:noisy_ablation} in appendix).

%% file: sections/analysis.tex
\section{Analysis}
\label{sec:analysis}
In light of the importance of BBF's components, we discuss possible consequences of our findings for other algorithms. %

\begin{figure}[t]
    \centering
    \includegraphics[width=\linewidth]{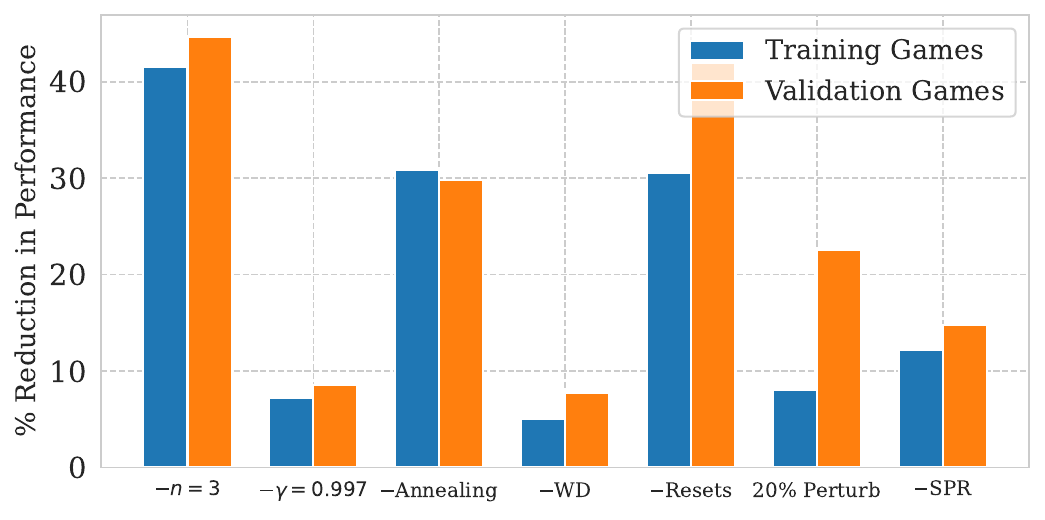}
    \vspace{-0.3cm}
    \caption{\textbf{Validating BBF design choices at RR=2 on 29 unseen games}. While Atari 100K training set consists of 26 games, we evaluate the performance of various components in BBF on 29 validation games in ALE that are not in Atari 100K. 
    Interestingly, all BBF components lead to a large performance improvement on unseen games.  Specifically, we measure the \% decrease in human-normalized IQM performance relative to the full BBF agent at RR=2. 
    } 
    \label{fig:ablation_robustness}
    \vspace{-0.4cm}
\end{figure}

\paragraph{The importance of self-supervision.} One unifying aspect of the methods compared in \autoref{fig:toplinePerformance} is that they all use some form of self-supervised objective. In sample-constrained scenarios, like the one considered here, relying on more than the temporal-difference backups is likely to improve learning  speed, provided the self-supervised losses are consistent with the task at hand. We test this by removing the SPR objective (inherited from SR-SPR) from BBF, and observe a substantial performance degredation (see \cref{fig:bbfAblations}).
It is worth noting that EfficientZero uses a self-supervised objective that is extremely similar to SPR, a striking commonality between BBF and EfficientZero.

\paragraph{Sample efficiency via more gradient steps.} The original DQN agent \citep{mnih2015human} has a replay ratio of $0.25$, which means a gradient update is performed only after every 4 environment steps. In low-data regimes, it is more beneficial to perform more gradient steps, although many algorithms cannot benefit from this without additional regularization~\citep{doro2023sampleefficient}. As \autoref{fig:replayRatioComparison} confirms, performance of BBF grows with increasing replay ratio in the same manner as its base algorithm, SR-SPR. More strikingly, we observe a \textit{linear} relationship between the performance of BBF and SR-SPR across all replay ratios, with BBF performing roughly 0.45 IQM above SR-SPR. While the direction of this relationship is intuitive given the network scaling introduced by BBF, its linearity is unexpected, and further investigation is needed to understand the nature of the interaction between replay ratio and network scaling.

One interesting comparison to note is that, although EfficientZero uses a replay ratio of 1.2, they train with a batch size that is 8 times larger than ours. Thus, their {\em effective} replay ratio is comparable to ours.

\begin{figure}[t]
    \centering
    \includegraphics[width=0.9\linewidth]{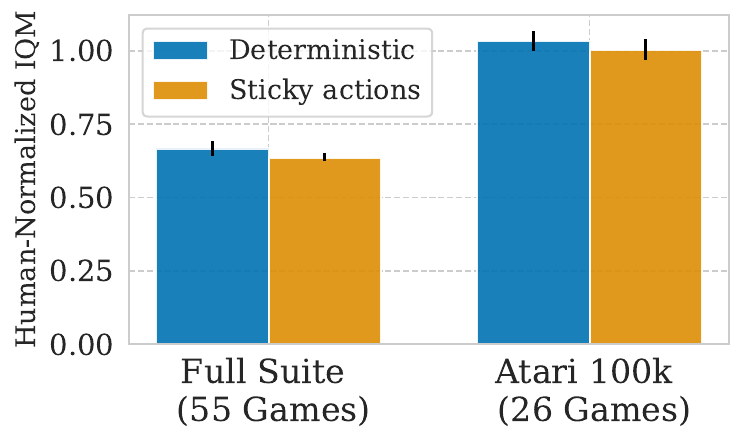}
    \vspace{-0.2cm}
    \caption{\textbf{Evaluating BBF on ALE with and w/o sticky actions}. We report IQM human-normalized performance at replay ratio 8 on 26 games in Atari 100K as well the full set of 55 games in ALE. While performance on the full set of 55 games is lower, neither setting has its performance significantly affected by sticky actions.
    } 
    \label{fig:55_sticky_actions}
    \vspace{-0.4cm}
\end{figure}

\paragraph{The surprising importance of target networks} Many prior works on Atari 100k, such as DrQ and SPR~\citep{kostrikov2020image, schwarzer2021dataefficient} chose not to use target networks, seeing them unnecessary or an impediment to sample efficiency. Later, \citet{doro2023sampleefficient} re-introduced an exponential moving average target network, used both for training and action selection, and found that it improved performance somewhat, especially at high replay ratios. With network scaling, however, using a target network becomes a critical, but easy-to-overlook, component of the algorithm at all replay ratios (see \autoref{fig:target_figure}).

\paragraph{Reset Strength}
Increasing the replay ratio is in general challenging, as explored by \citet{fedus20revisiting} and \citet{kumar2020implicit}. Periodic resetting, as suggested by \citet{nikishin22primacy} and \citet{doro2023sampleefficient}, has proven effective to enable scaling to larger replay ratios, quite possibly a result of reduced overfitting. This is confirmed in~\autoref{fig:bbfAblations}, where the importance of resets is clear. Further, \autoref{fig:bbfAblations} and \autoref{fig:ablation_robustness} demonstrate the added benefit of more aggressive perturbations, relative to SR-SPR.

\paragraph{Scale is not enough on its own.} The na{\" i}ve approach of simply scaling the capacity of the CNN used by SR-SPR turns out to be insufficient to improve performance. Instead, as Figure~\ref{fig:scalingWidths} shows, the performance of SR-SPR collapses as network size increases. As discussed in \autoref{sec:method}, it is interesting to observe that the smaller Impala-CNN ResNet (as measured by number of parameters and FLOPs) yields stronger performance at all width scales.

\paragraph{Computational efficiency.} As machine learning methods become more sophisticated, an often overlooked metric is their computational efficiency. Although EfficientZero trains in around 8.5 hours, it requires about 512 CPU cores and 4 distributed GPUs. IRIS uses half of an A100 GPU for a week per run. SR-SPR, at its highest replay ratio of 16, uses 25\% of an A100 GPU and a single CPU for roughly 24 hours. Our BBF agent at replay ratio 8 takes only 10 hours with a single CPU and half of an A100 GPU. Thus, measured by GPU-hours, BBF provides the best trade-off between performance and compute (see \autoref{fig:toplinePerformance}).

%% file: sections/discussion.tex
\section{Revisiting the Atari 100k benchmark}
\label{sec:discussion}

\begin{figure}[!t]
    \centering
    \includegraphics[width=\linewidth]{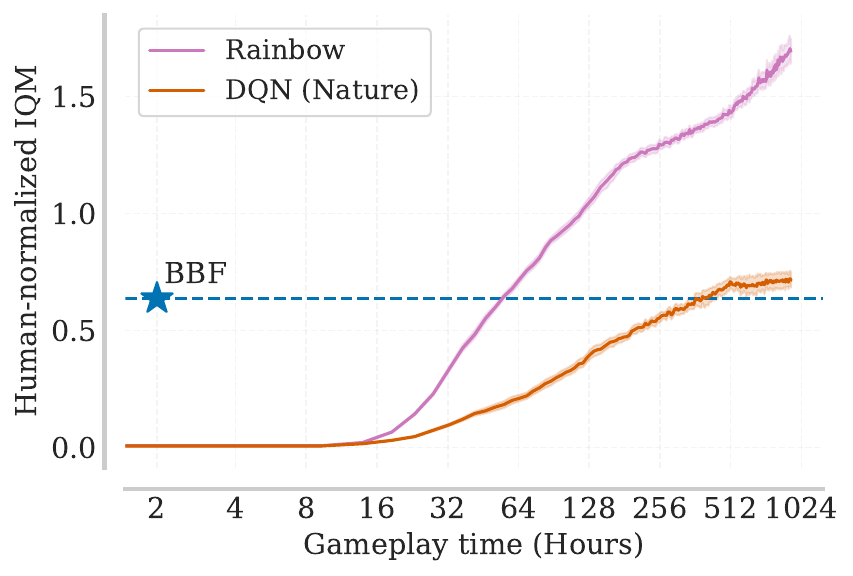}
    \vspace{-0.3cm}
    \caption{\textbf{Sample efficiency progress on ALE}, measured via human-normalized IQM over 55 Atari games with sticky actions, as a function of amount of human game play hours, with BBF at RR=8. Shaded regions show 95\% CIs.} 
    \label{fig:gameplay_efficiency_progress}
    \vspace{-0.5cm}
\end{figure}

A natural question is whether there is any value in continuing to use the Atari 100K benchmark, given that both EfficientZero and BBF are able to achieve human-level performance (IQM $\geq 1.0$) in just 100K steps. When considering this, it is important to remember that IQM is an {\em aggregate} measure. Indeed, in the left panel of \autoref{fig:optimalitygap} we can see there is still room for improvement with regards to the {\em optimality gap}, which measures the amount by which each
algorithm fails to meet a minimum score of $1.0$ \citep{agarwal2021deep}. Specifically, despite monotonic progress over the years, no agent is yet able to achieve human-level performance on all 26 games, which would yield an optimality gap of zero, without using dramatically more than two hours of data~\citep{kapturowski2022human}.

\textbf{Overfitting on Atari 100K}. Another important consideration is that the Atari 100K benchmark uses only 26 of the 55 games from the full ALE suite, and it does not include sticky actions\footnote{With 25\% probability, the environment will execute the previous action again, instead of the agent's executed action.} \citep{machado18revisiting}, which may make tasks significantly harder. Since we extensively benchmark BBF on Atari 100K, this raises the question of whether BBF works well on unseen Atari games and with sticky actions.

\begin{figure}[t]
    \centering
    \includegraphics[width=\linewidth]{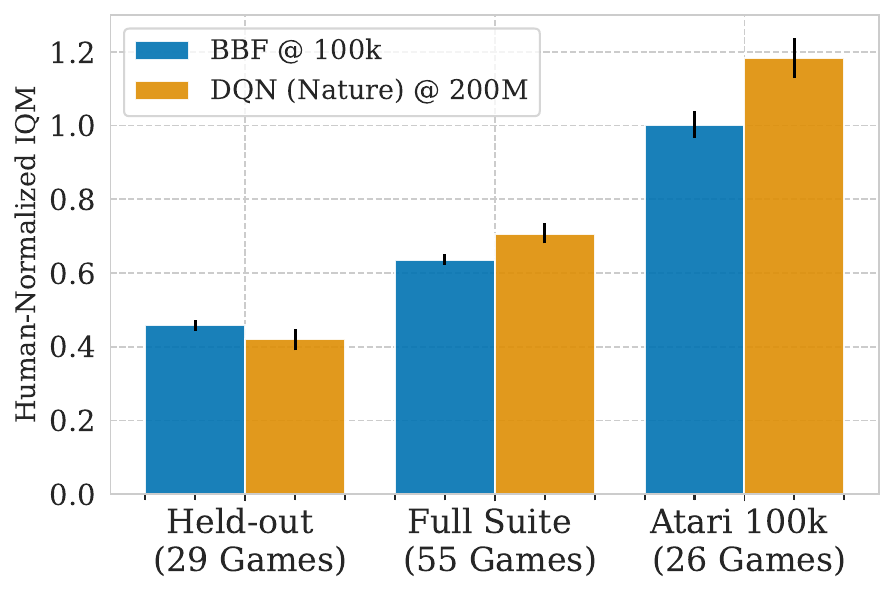}
    \vspace{-0.3cm}
    \caption{\textbf{Comparing performance on the 29 unseen games to the 26 Atari 100k games}. BBF trained with sticky actions at RR=8 for 100k steps approximately matches DQN (Nature) with 500 times more training data on each set. While we find that the 29 games not included in the Atari 100k setting are significantly harder than the 26 Atari 100k games, we see no evidence that BBF has overfitted to Atari 100k compared to DQN.
    } 
    \label{fig:game_group_comparison}
    \vspace{-0.4cm}
\end{figure}

Fortunately, it does. In \autoref{fig:55_sticky_actions}, we compare the performance of BBF on all 55 games with sticky actions, and show that sticky action do not significantly harm performance. We do observe that the held-out games not included in the Atari 100k set are significantly more challenging than the 26 Atari 100k games (see \autoref{fig:game_group_comparison}) -- but this is even more true for baselines such as DQN (Nature) that did not use Atari 100k. Furthermore, as shown in \autoref{fig:ablation_robustness}, we find that BBF's design choices generally provide even more benefit on these held-out games, possibly due to their increased difficulty.

\begin{figure*}[!t]
    \centering
    \includegraphics[width=0.95\linewidth]{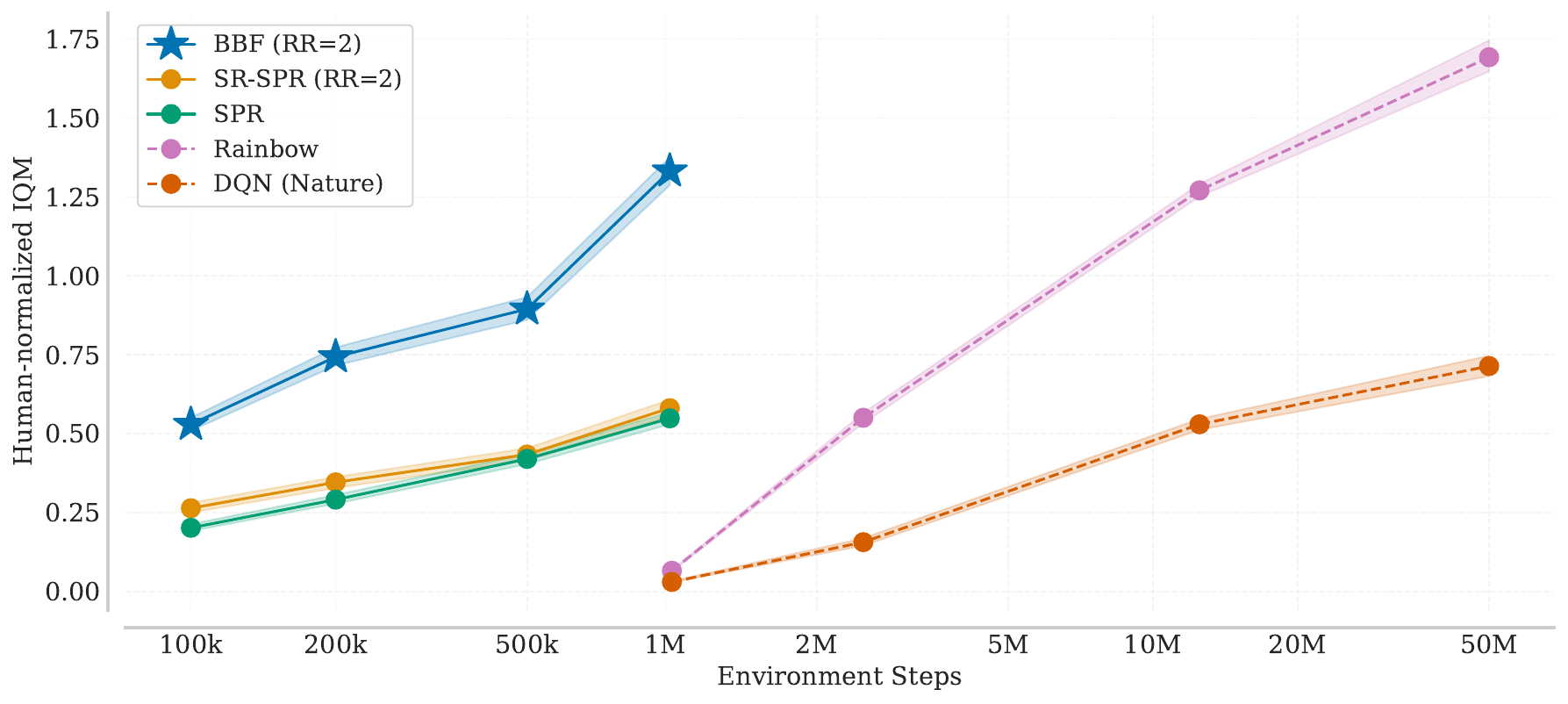}
    \vspace{-0.3cm}
    \caption{\textbf{Learning curves for BBF, SR-SPR and SPR at replay ratio 2}, measured via human-normalized IQM over 55 Atari games with sticky actions, as a function of number of environment interactions. Shaded regions show 95\% CIs.} 
    \label{fig:data_scaling}
    \vspace{-0.2cm}
\end{figure*}

\paragraph{New Frontiers}
In fact, BBF works so well on the standard Atari setting that it is able to roughly match DQN's performance at 256 hours with only two hours of gameplay time (\autoref{fig:gameplay_efficiency_progress}). This suggests a clear new milestone for the community: can we match Rainbow's final performance with just two hours of gameplay? To facilitate future research toward this, we release scores on the set of 55 games with sticky actions, at various scales and replay ratios.

\paragraph{Data Scaling} Prior works have indicated that many sample-efficient RL algorithms plateau in performance when trained for longer than they were originally designed for~\citep[e.g.,][]{agarwalreincarnating}. To examine this phenomenon, we train BBF, SPR and SR-SPR at replay ratio 2 out to one million environment steps (\autoref{fig:data_scaling}), keeping all parameters unchanged (including conducting resets as normal past 100k steps). We observe that SPR and SR-SPR experience stagnating performance, with SR-SPR's advantage over SPR fading by 1M steps. BBF, however, remains consistently ahead of both, matching DQN's final performance before 200k environment steps and matching Rainbow's performance at 20M environment steps by 1M steps. We note that this experiment costs only 2.5 times more than training at replay ratio 8 to 100k steps, so we encourage other researchers to run similar experiments.

\begin{figure}[hb]
    \centering
    \includegraphics[width=\columnwidth]{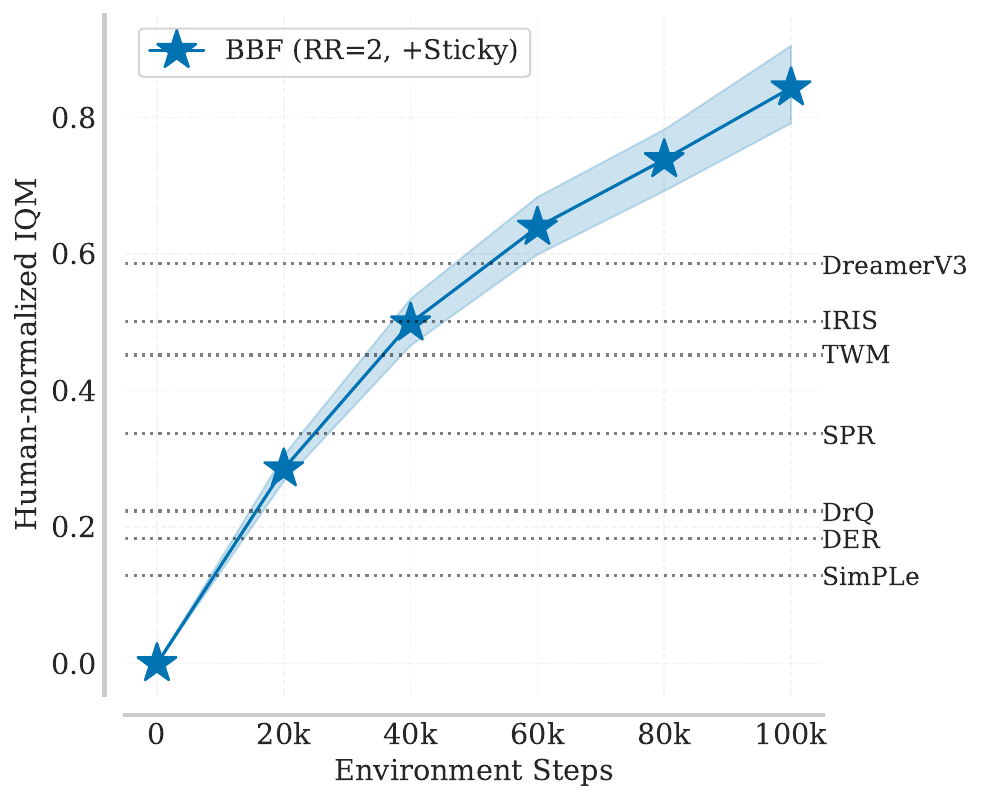}
    \caption{\textbf{IQM Human-normalized learning curve for BBF at RR=2 with sticky actions on the 26 Atari 100k games}, with final performances of many recent algorithms after they have trained for 100k steps. Even a weakened BBF outperforms all by 50k steps.}
    \label{fig:extreme_sample_efficiency}
    \vspace{-1cm}
\end{figure}

Additionally, we note in \autoref{fig:extreme_sample_efficiency} that it is possible to compare algorithms even with extremely small amounts of data, such as 20k or 50k steps, by which point BBF at replay ratio 2 (even with sticky actions enabled) outperforms most recently proposed algorithms \citep{robine2023transformerbased, micheli2023transformers, danijar_dreamerv3}, which did not use sticky actions. We thus suggest that compute-constrained groups consider this setting, as training BBF at replay ratio 2 for 40k environment steps takes only half of an A100 for 1 hour.

\section{Discussion and Future Work}

We introduced BBF, an algorithm that is able to achieve super-human level performance on the ALE with only 2-hours of gameplay. Although BBF is not the first to achieve this milestone, it is able to do so in a computationally efficient manner. Furthermore, BBF is able to better handle the scaling of networks and replay ratios, which are crucial for network expressivity and learning efficiency. Indeed, \autoref{fig:scalingWidths} suggests that BFF is better-able to use over-parameterized networks than prior agents.

The techniques necessary to achieve this result invite a number of research questions for future work. Large replay ratios are a key element of BFF's performance, and the ability to scale them is due to the periodic resets incorporated into the algorithm. These resets are likely striking a favourable balance between catastrophic forgetting and network plasticity. An interesting avenue for future research is whether there are other mechanisms for striking this balance that perhaps are more targeted (e.g. not requiring resetting the full network, as was recently explored by \citet{sokar23redo}). We remarked on the fact that all the methods compared in \autoref{fig:toplinePerformance} use a form of self-supervision.  Would other self-supervised losses (e.g. \citep{mazoure2020deep,castro2021mico, agarwal2021contrastive}) produce similar results? Surprisingly, \citet{li2022does} argue that self-supervision from pixels does not improve performance; our results seem to contradict this finding.

Recent attention has shifted towards more realistic benchmarks~\citep{fanminedojo} but such benchmarks exclude the majority of researchers outside certain resource-rich labs, and may require an alternative paradigm~\citep{agarwalreincarnating}. One advantage of the Atari 100k benchmark is that, while still a challenging benchmark, it is relatively cheap compared to other benchmarks of similar complexity. However, despite its apparent saturation, scientific progress can still be made on this benchmark if we expand its scope. We hope our work provides a solid starting point for this.

Overall, we hope that our work inspires other researchers to continue pushing the frontier of sample efficiency in deep RL forward, to ultimately reach human-level performance across all tasks with human-level or superhuman efficiency.

\clearpage

%% file: sections/appendix.tex
\section{Additional Results}
\label{sec:additional_results}

\begin{table}[ht]
    \centering
    \footnotesize
\begin{tabular}{llllllllll}
\toprule
{} &   Random &    Human &      DER & DrQ($\epsilon$) &      SPR &     IRIS &    SR-SPR & EfficientZero &      BBF \\
\midrule
Alien  &  227.8  &  7127.7  &  802.3  &  865.2  &  841.9  &  420.0  &  1107.8  &  808.5  &  \textbf{1173.2}  \\
Amidar  &  5.8  &  1719.5  &  125.9  &  137.8  &  179.7  &  143.0  &  203.4  &  148.6  &  \textbf{244.6}  \\
Assault  &  222.4  &  742.0  &  561.5  &  579.6  &  565.6  &  1524.4  &  1088.9  &  1263.1  &  \textbf{2098.5}  \\
Asterix  &  210.0  &  8503.3  &  535.4  &  763.6  &  962.5  &  853.6  &  903.1  &  \textbf{25557.8}  &  3946.1  \\
BankHeist  &  14.2  &  753.1  &  185.5  &  232.9  &  345.4  &  53.1  &  531.7  &  351.0  &  \textbf{732.9}  \\
BattleZone  &  2360.0  &  37187.5  &  8977.0  &  10165.3  &  14834.1  &  13074.0  &  17671.0  &  13871.2  &  \textbf{24459.8}  \\
Boxing  &  0.1  &  12.1  &  -0.3  &  9.0  &  35.7  &  70.1  &  45.8  &  52.7  &  \textbf{85.8}  \\
Breakout  &  1.7  &  30.5  &  9.2  &  19.8  &  19.6  &  83.7  &  25.5  &  \textbf{414.1}  &  370.6  \\
ChopperCommand  &  811.0  &  7387.8  &  925.9  &  844.6  &  946.3  &  1565.0  &  2362.1  &  1117.3  &  \textbf{7549.3}  \\
CrazyClimber  &  10780.5  &  35829.4  &  34508.6  &  21539.0  &  36700.5  &  59324.2  &  45544.1  &  \textbf{83940.2}  &  58431.8  \\
DemonAttack  &  152.1  &  1971.0  &  627.6  &  1321.5  &  517.6  &  2034.4  &  2814.4  &  13003.9  &  \textbf{13341.4}  \\
Freeway  &  0.0  &  29.6  &  20.9  &  20.3  &  19.3  &  \textbf{31.1}  &  25.4  &  21.8  &  25.5  \\
Frostbite  &  65.2  &  4334.7  &  871.0  &  1014.2  &  1170.7  &  259.1  &  \textbf{2584.8}  &  296.3  &  2384.8  \\
Gopher  &  257.6  &  2412.5  &  467.0  &  621.6  &  660.6  &  2236.1  &  712.4  &  \textbf{3260.3}  &  1331.2  \\
Hero  &  1027.0  &  30826.4  &  6226.0  &  4167.9  &  5858.6  &  7037.4  &  8524.0  &  \textbf{9315.9}  &  7818.6  \\
Jamesbond  &  29.0  &  302.8  &  275.7  &  349.1  &  366.5  &  462.7  &  389.1  &  517.0  &  \textbf{1129.6}  \\
Kangaroo  &  52.0  &  3035.0  &  581.7  &  1088.4  &  3617.4  &  838.2  &  3631.7  &  724.1  &  \textbf{6614.7}  \\
Krull  &  1598.0  &  2665.5  &  3256.9  &  4402.1  &  3681.6  &  6616.4  &  5911.8  &  5663.3  &  \textbf{8223.4}  \\
KungFuMaster  &  258.5  &  22736.3  &  6580.1  &  11467.4  &  14783.2  &  21759.8  &  18649.4  &  \textbf{30944.8}  &  18991.7  \\
MsPacman  &  307.3  &  6951.6  &  1187.4  &  1218.1  &  1318.4  &  999.1  &  1574.1  &  1281.2  &  \textbf{2008.3}  \\
Pong  &  -20.7  &  14.6  &  -9.7  &  -9.1  &  -5.4  &  14.6  &  2.9  &  \textbf{20.1}  &  16.7  \\
PrivateEye  &  24.9  &  69571.3  &  72.8  &  3.5  &  86.0  &  \textbf{100.0}  &  97.9  &  96.7  &  40.5  \\
Qbert  &  163.9  &  13455.0  &  1773.5  &  1810.7  &  866.3  &  745.7  &  4044.1  &  \textbf{14448.5}  &  4447.1  \\
Roadrunner  &  11.5  &  7845.0  &  11843.4  &  11211.4  &  12213.1  &  9614.6  &  13463.4  &  17751.3  &  \textbf{33426.8}  \\
Seaquest  &  68.4  &  42054.7  &  304.6  &  352.3  &  558.1  &  661.3  &  819.0  &  1100.2  &  \textbf{1232.5}  \\
UpNDown  &  533.4  &  11693.2  &  3075.0  &  4324.5  &  10859.2  &  3546.2  &  \textbf{112450.3}  &  17264.2  &  12101.7  \\
\midrule
Games $>$ Human  &  0  &  0  &  2  &  3  &  6  &  9  &  9  &  \textbf{14}  &  12  \\
IQM ($\uparrow$)  &  0.000  &  1.000  &  0.183  &  0.280  &  0.337  &  0.501  &  0.631  &  1.020  &  \textbf{1.045}  \\
Optimality Gap ($\downarrow$)  &  1.000  &  0.000  &  0.698  &  0.631  &  0.577  &  0.512  &  0.433  &  0.371  &  \textbf{0.344}  \\
Median ($\uparrow$)  &  0.000  &  1.000  &  0.189  &  0.313  &  0.396  &  0.289  &  0.685  &  \textbf{1.116}  &  0.917  \\
Mean ($\uparrow$)  &  0.000  &  1.000  &  0.350  &  0.465  &  0.616  &  1.046  &  1.272  &  1.945  &  \textbf{2.247}  \\
\bottomrule
\end{tabular}

    \caption{\textbf{Scores and aggregate metrics for BBF and competing methods across the 26 Atari 100k games}. Scores are averaged across 50 seeds per game for BBF, 30 for SR-SPR, 5 for IRIS, 3 for EfficientZero, and 100 for others.}
    \label{tab:my_label}
\end{table}

\begin{figure*}[b]
    \centering
    \includegraphics[width=\textwidth]{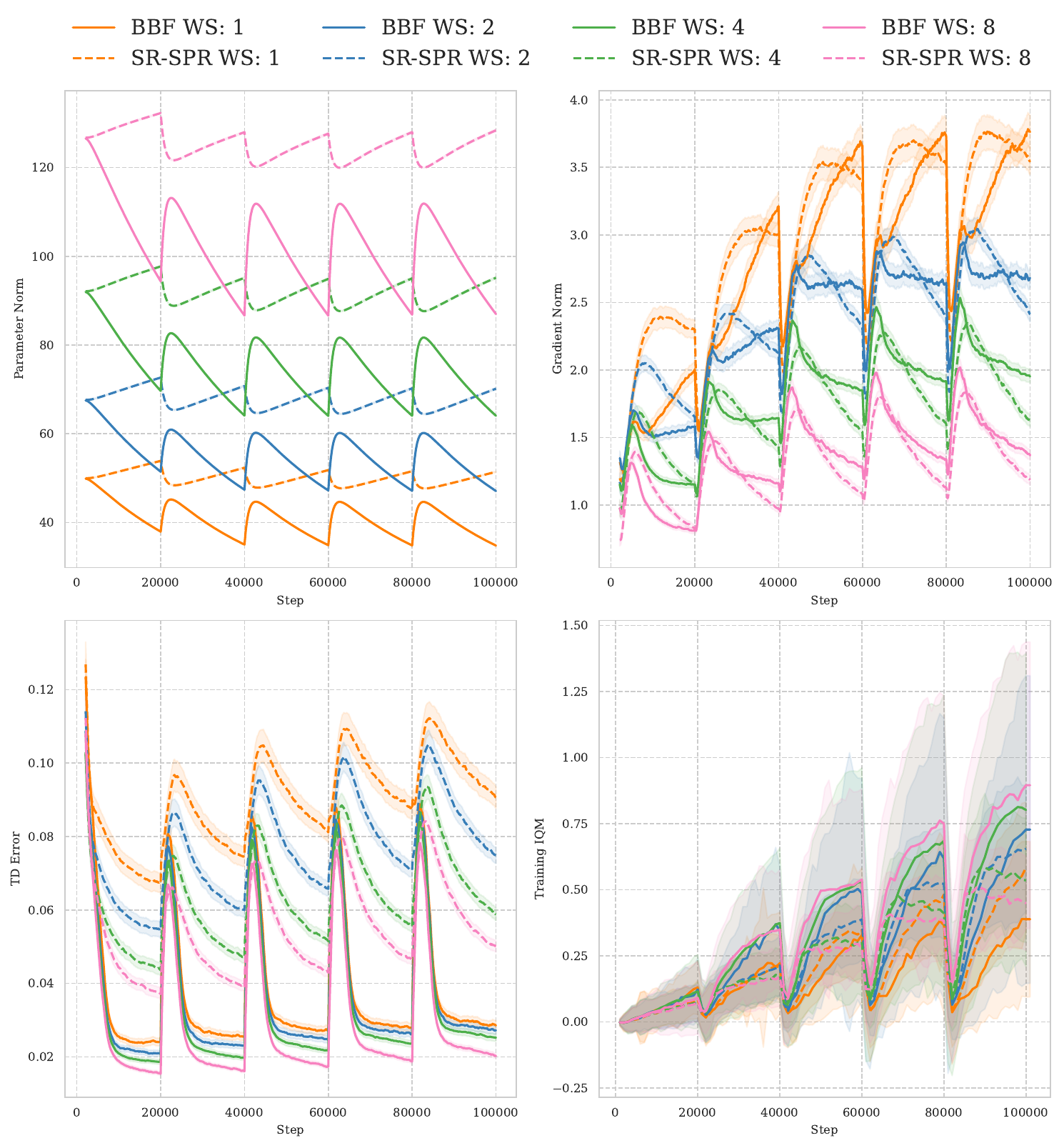}
    \caption{\textbf{Learning curves for BBF and SR-SPR at RR=2 with a ResNet encoder at various width scales}, on the 26 Atari 100k games. Larger networks consistently have lower TD errors and higher gradient norms, and higher parameter norms, but only BBF translates this to higher environment returns. The large, systematic difference in TD error between BBF and SR-SPR is due to BBF's use of a shorter update horizon, which makes each step of the TD backup easier to predict.}
    \label{fig:learning_curves}
\end{figure*}

\begin{figure}
    \centering
    \includegraphics[width=\columnwidth]{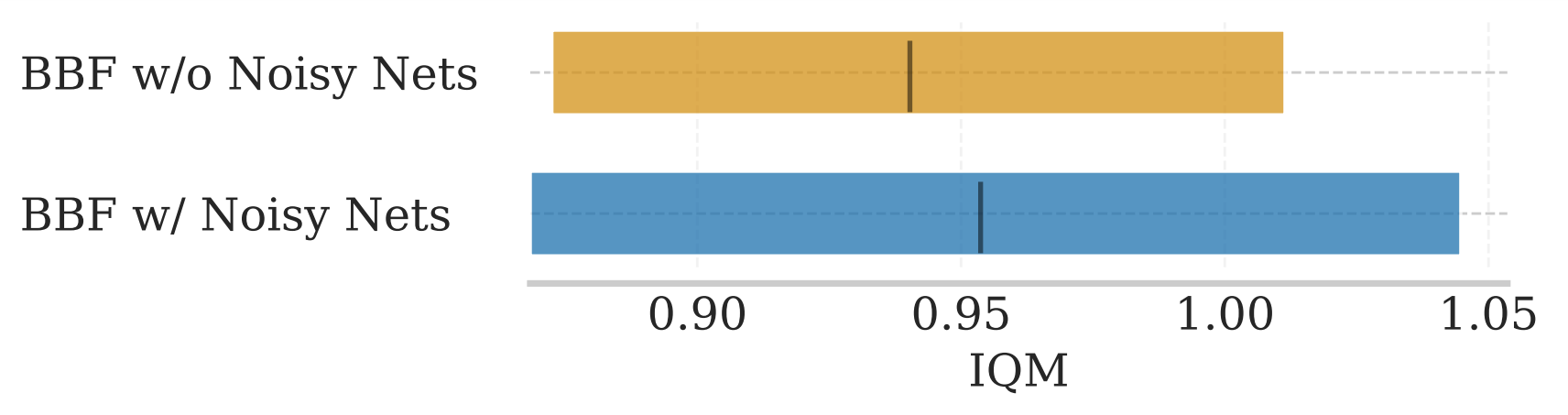}
    \caption{BBF at RR=2 on the 26 Atari 100k tasks, with and without Noisy Nets.}
    \label{fig:noisy_ablation}
\end{figure}